# Towards the Long-Endurance Flight of an Insect-Inspired, Tailless, Two-Winged, Flapping-Wing Flying Robot

Hoang Vu Phan, Steven Aurecianus, Thi Kim Loan Au, Taesam Kang, and Hoon Cheol Park

*Abstract*—A hover-capable insect-inspired flying robot that can remain long in the air has shown its potential use for both confined indoor and outdoor applications to complete assigned tasks. In this letter, we report improvements in the flight endurance of our 15.8 g robot, named KUBeetle-S, using a low-voltage power source. The robot is equipped with a simple but effective control mechanism that can modulate the stroke plane for attitude stabilization and control. Due to the demand for extended flight, we performed a series of experiments on the lift generation and power requirement of the robot with different stroke amplitudes and wing areas. We show that a larger wing with less inboard wing area improves the lift-to-power ratio and produces a peak lift-to-weight ratio of 1.34 at 3.7 V application. Flight tests show that the robot employing the selected wing could hover for 8.8 minutes. Moreover, the robot could perform maneuvers in any direction, fly outdoors, and carry payload, demonstrating its ability to enter the next phase of autonomous flight.

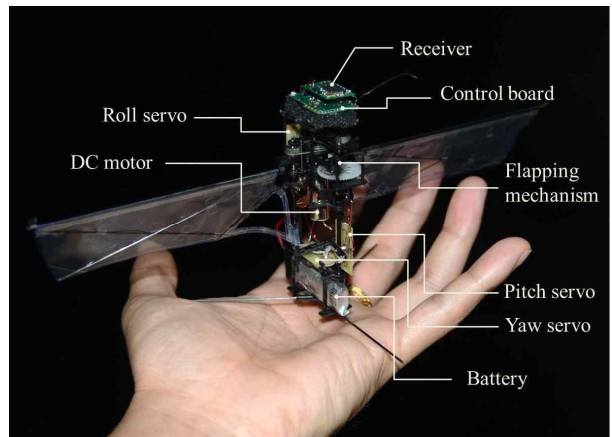

Fig. 1. The insect-inspired KUBeetle-S robot that can perform free controlled flight.

## I. INTRODUCTION

INSECTS in nature rely on their flapping wings to perform agile flight capabilities from hovering to evasive takeoffs [1], and rapid banked turns [2]. These abilities have inspired many researchers to build an insect-inspired, tailless, hover-capable, flapping-wing robots, not only for discovering the secrets of natural flight but also for useful applications, such as entering the confined spaces of collapsed buildings, environmental monitoring, or even for entertainment as human-friendly toys. Moreover, at the smaller scale of insects, a flapping wing, based on its alternative unsteady aerodynamics, generates a higher propulsive efficiency than a conventional rotary wing [3]. However, building such a small scale robot with a body mass of less than a few grams is technically challenging and requires nontraditional actuators other than electromagnetic motors, which are relatively inefficient and hard to fabricate at a small scale [3]. Therefore, oscillating piezoelectric actuators have been selected due to their light-weight, high power density and low power consumption [4]–[6], while no need of conversion linkages as found in motor-driven systems. Nevertheless, the actuators require applied voltage of hundreds of volts that need additional onboard power electronics to create high-voltage drive signals from low-voltage power supplies. As fruits of long-term efforts, the insect-scale, piezo-driven RoboBee could perform stable controlled flights with off-board components [4], or uncontrolled takeoff with onboard power source and electronics [7]. However, the robot still requires further substantial improvements for implementation of all onboard components, to allow efficient and stable free flight.

At larger scales, such as that of large insects and giant hummingbirds (~ 7–20 g), several insect-inspired, motor-driven, two-winged robots have been developed and successfully flown [8]–[12]. However, most of them generate inefficient propulsion, which causes overheating of the motors to stay airborne for several tens of seconds [10], [11], or still require off-board power sources [12]. The 19 g Nano Hummingbird is the first two-winged robot to perform a free controlled flight with all onboard components, and still remain the top position in flight endurances of 4 minutes and 11

Manuscript received February 24, 2020; Revised May 13, 2020; Accepted June 16, 2020.

This paper was recommended for publication by Editor Xinyu Liu upon evaluation of the associate editor and reviewers' comments.

H. V. Phan is supported by the KU Brain Pool Program of Konkuk University, South Korea.

H. V. Phan and T. K. L. Au are with the Department of Smart Vehicle Engineering, Konkuk University, Seoul 05029, South Korea (e-mail: vu113@konkuk.ac.kr; aukimloan@gmail.com).

S. Aurecianus and T. Kang are with the Department of Aerospace Information Engineering, Konkuk University, Seoul 05029, South Korea (e-mail: steven.aurecianus@gmail.com; tskang@konkuk.ac.kr).

H. C. Park is with the Department of Smart Vehicle Engineering, Konkuk University, Seoul 05029, South Korea (corresponding author, e-mail: hcpark@konkuk.ac.kr).

Digital Object Identifier (DOI): see top of this page.

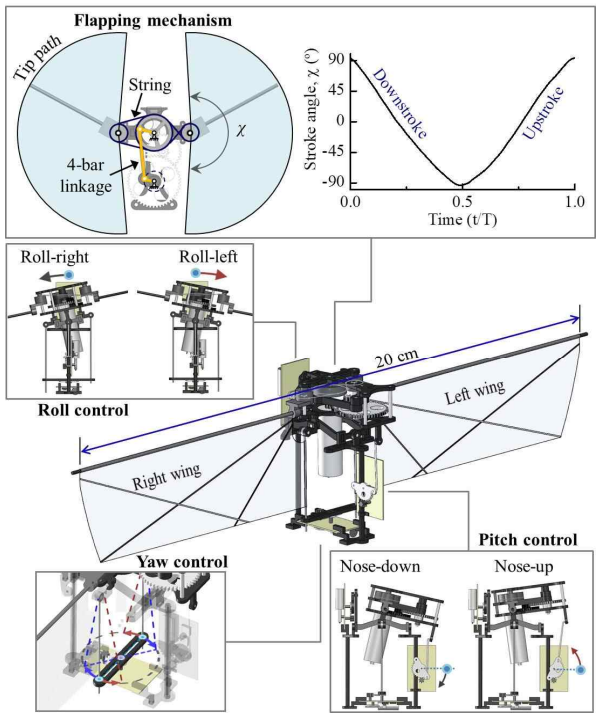

Fig. 2. Flapping and control mechanisms used in the KUBeetle-S robot. Two wings are driven by a coreless DC motor (Chaoli CL720) through a gear box of 28:1 to amplify the motor torque.

minutes (Saturn version) with and without carrying payload, respectively [9]. This demonstrates its availability for applications. However, the Nano Hummingbird's lack of documentation makes it difficult for academic researchers to adapt the design and fabricate a similar one. To efficiently remain airborne, an alternative design that uses four clap-and-fling wings has been proposed [13], [14]. However, it is not considered a truly biomimetic configuration.

Recent achievement in stable flight of a two-winged, tailless, hover-capable robot is an important step forward. However, how long it can stably fly should be considered as a prerequisite for applications to complete assigned tasks. Therefore, numerous studies have been conducted to improve the performance of the flapping wing [15]–[17]. For example, an experimental study by Nan et al. [15] found that a trapezoidal wing with an aspect ratio similar to that of a real hummingbird wing provides better lift-to-power ratio. Effects of wing morphological and inertial parameters were also investigated in [16]. On the other hand, to save weight and minimize inertial power, flapping-wing robots based on direct-driven actuation incorporating compliant wings were proposed [12], [18]. However, due to size and weight constraints, the recently released 20.4 g direct-driven robot [19] (as well as two-winged robots other than the Nano Hummingbird) still requires a high voltage power source to lift off and sustain flight of only several tens of seconds. We also successfully demonstrated stable flight of our tailless two-winged robot, named KUBeetle-S [20]. However, its maximum flight endurance was limited to less than 3 minutes due to overheating of the coreless DC motor at a 7.4 V application.

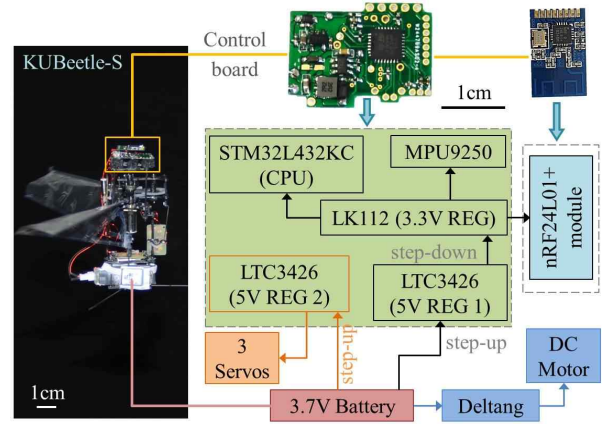

Fig. 3. Custom-built control board and power schematics used in the KUBeetle-S robot.

Longer flight caused degradation and failure of the driving motor.

In this letter, we report how the flight endurance of our robot was improved by maximizing the lift-to-power ratio with a low-voltage power source (3.7 V), which is within the operating range of the selected motor (Fig. 1). To this end, we experimentally investigate several wing configurations with different stroke amplitudes and wing areas. The body mass of the robot is also reduced by using lightweight sub-micro servos, making it the lightest two-winged robot so far that can sustain free controlled flight. For heading stabilization and control, we use the angular rate signal from an onboard gyroscope, instead of the signal from an external magnetometer board in our previous robot version [20]. Additionally, unlike the Nano Hummingbird, our robot is built using commercially available components, except for the control board, which is also appropriately redesigned. Therefore, any academic researcher can attempt to build a similar insect-inspired flying robot.

This report is organized as follows. Section II introduces the design and prototype of the flapping-wing and control mechanisms, and attitude stabilization system. Section III presents the experimental tests on lift generation and power requirement for wings with different stroke amplitudes and areas to find a proper wing that provides the highest lift-to-power ratio. Section IV then describes flight tests of the robot. Finally, Section V provides the discussion of our work, while Section VI concludes the paper.

## II. KUBEETLE-S

### A. Flapping-wing mechanism

We briefly summarize the flapping-wing mechanism design as [11] presented its detail. To obtain a high stroke amplitude of about 190° that mimics the wing motion of the horned beetle, *Allomyrina dichotoma* [21], we utilized a combination of four-bar linkage (to convert rotary motion of the DC motor to flapping motion of the large pulley) and pulley–string mechanism (to amplify the flapping motion) (Fig. 2) [10]. The string is twisted on one side to create symmetric motions of the left and right wings. This high amplitude allows the presence of near-clap-and-fling effects at both dorsal and ventral stroke reversals to enhance lift generation [22].

We used 0.8 mm thick carbon/epoxy panels to fabricate the frames of the flapping mechanism. A wing membrane made of 10 μm Mylar film was reinforced by 0.3 mm carbon rods as veins. The arrangement of the veins was previously investigated to obtain a high lift-to-power ratio [23]. Along the leading edge and wing root margin, the membrane was fabricated as a sleeve for free rotation during flapping motion, creating wing deformation to generate useful forces [9], [10].

### B. Control mechanism

Without tail stabilizations, the robot relies on its wing kinematics modulation to generate control torques for attitude stabilization and control. For generation of pitch and roll torques, the robot tilts its wing stroke planes of both the left and right wings, resulting in change in the direction of the resultant lift (Fig. 2) [20]. Due to the constrained wing–root spars, tilting the stroke plane also results in the modulation of wing twist to produce additional control torques, allowing fast response of the robot's attitude to the pitch and roll control signals. For yaw (heading) control, the wing–root spars of the left and right wings are adjusted in opposite directions, resulting in asymmetric horizontal drags in the two wings to produce yaw torque. The simplicity of the mechanism allows the use of three lightweight sub-micro servos (0.6 g LZ servos) to save weight.

### C. Attitude stabilization system

Inherent flight instability of the tailless robot requires an active feedback control system to maintain the robot stability when airborne. To this end, we built a four-layer control board containing an ARM 32-bit Cortex-M4 (STM32L432KC) microprocessor, a 9-axis IMU (MPU9250) with 3-axis gyroscope, accelerometer and magnetometer, and two step-up (LTC3426) and one step-down (LK112) power regulators (Fig. 3). Even though the magnetometer is equipped, it was not activated in this work. The board was placed on a damping foam at the top of the robot to reduce mechanically-produced high noise caused by the flapping motion (Fig. 1). A 3.7 V (160 mAh, 25 C) single lithium polymer battery was used as the power source of the robot to directly power a receiver and driving motor. Meanwhile, it provides a 3.3 V voltage via two regulators (1 step-up and 1 step-down regulator) to power the control board, and 5 V voltage through a step-up regulator to power the three sub-micro servos (Fig. 3). We used an off-the-shelf, 4-channel, 2.4 GHz DSM2 compatible Deltang DT-Rx35 receiver to communicate with the pilot. The lightweight receiver is equipped with an onboard brushed electronic speed controller (ESC) via a field-effect transistor (FET) with a conservative 2A rating and a 3.0 V Low Voltage Cutoff (LVC). To acquire real time flight data, the robot was additionally equipped with a 2.4 GHz nRF24L01+ transceiver module transmitting at a data rate of approximately 100 Hz.

For stabilization, we used a feedback controller with proportional (P) and derivative (D) terms to sense the attitudes and angular rates of the robot via an in-house code developed in an Mbed interface (version 1.10.25.0). For pitch and roll stabilization, we used signals from both accelerometer (providing roll ($\phi$), pitch ($\theta$), and yaw ($\psi$) attitude angles) and gyroscope (providing roll ($\omega_r$), pitch ($\omega_p$), and yaw ($\omega_y$) angular rates) via a combination of low-pass and Kalman filters [24] to obtain angle estimation for the P term. Meanwhile, the low-pass filtered pitch and roll angular rates were used for damping (D term).

For yaw angle estimation, our previous work utilized signals from the magnetometer [20]. However, this method requires a continuously updated calibration when the robot changes its location, which causes adjustments in the magnetic field. To avoid this issue in this work, yaw stabilization was obtained by using the angular rate signal from the gyroscope only. We integrated the low-pass filtered yaw rate to get yaw angle estimation for the P term, and used a low-pass filtered yaw rate for the D term, as follows:

$$y_y = k_{py}\left(\psi_d - \psi^g\right) + k_{dy}\left(\omega_{y,d} - \omega_y^f\right), \quad (1)$$

where, $y_y$ is the yaw control output; $k_{py}$ and $k_{dy}$ are the proportional and derivative gains, respectively; $\psi_d$ and $\omega_{y,d}$ are the desired states; $\omega_y^f$ is the actual filtered yaw rate; and $\psi^g$ can be obtained by using the following equation:

$$\psi_t^g = \psi_{t-1}^g + \omega_y^f \Delta t, \quad (2)$$

where, $t$ and $t-1$ denote the current and previous time steps, respectively, and $\Delta t$ is the time interval.

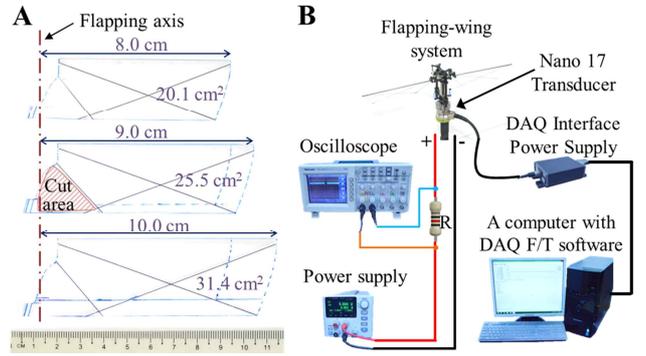

Fig. 4. Experimental setup. (A) Wings with different wing areas at the same aspect ratio of 3.2. (B) Experimental apparatus for force and power measurements.

## III. EXPERIMENTAL INVESTIGATION ON LIFT AND POWER

### A. Experimental setup

From the theoretical study on beetle flight, we found that beetles benefit from their high stroke amplitude of more than 180° to generate sufficient lift overcoming body mass and improve flight efficiency [25]. The study also showed that the beetle using a wing with larger area generates lift more efficiently than that with a smaller one. In this work, to confirm the finding, and to improve the flight efficiency and extend the flight time of the flapping-wing robot, we performed a series of experiments on lift generation and power requirement with different stroke amplitudes and wing areas (Fig. 4). We used an external power supply (E36103A,

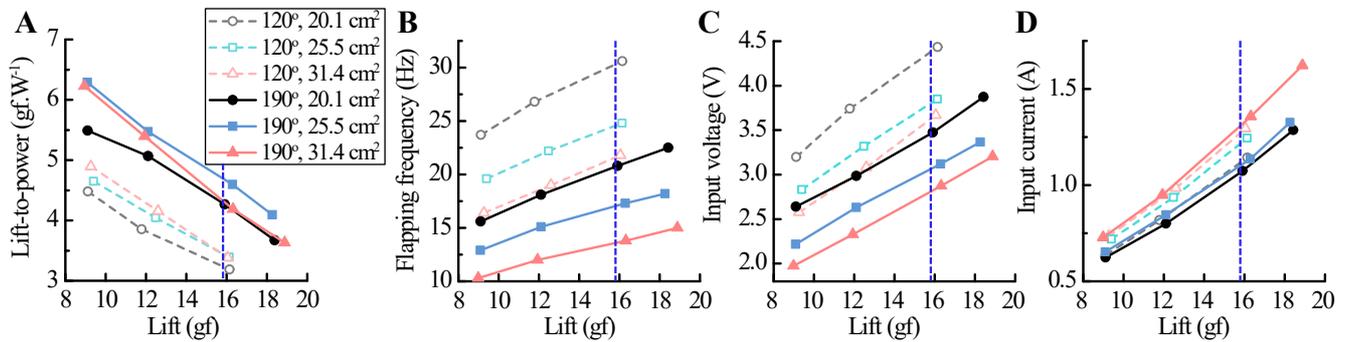

Fig. 5. Generation of lift or vertical force at different stroke amplitudes and wing areas. (A) Lift-to-power with respect to lift generation. (B) Lift versus flapping frequency (C) Lift versus input voltage. (D) Lift versus input current. Vertical dashed-blue line represents the body weight of the robot. Maximum force SD = ±0.2 gram force (gf). (1.0 gf = 9.8 mN).

Keysight) to excite the wings. A 6-axis load cell (Nano 17, ATI Industrial Automation, Inc., force resolution ≈ 0.32 gf) was used to measure the vertical force or lift generation of the robot at different stroke amplitudes of 120°, which were also used in other flapping wing robots [4], [26] and 190°, and for different wing areas of 20.1 cm$^2$, 25.5 cm$^2$ and 31.4 cm$^2$ (Fig. 4A). These wings have the same aspect ratio of 3.2, which is similar to that of the beetle's hindwing [25]. The flapping-wing system was mounted on the Nano 17 transducer through an adapter made of carbon pipe. Signals from the transducer were transmitted to a computer via a data acquisition (DAQ) system and converted to force data by using the installed ATI DAQ F/T software (ATI Industrial Automation, Version 1.0.4.2.2) and calibration data. The sampling frequency was set to 3.2 kHz to ensure accurate measurement in each flapping cycle. We measured the force by activating the load cell for approximately 3 s as idle conditions before and after exciting the flapping wings for approximately 4 s. The measurement is successfully acquired if the deviation between the average values of these idle conditions is less than 50% of the force resolution (0.32 gf) of the load cell. The cycle-average lift was obtained over 3 s flapping motions (30–90 flapping cycles at a frequency range of 10–30 Hz), excluding the data at the transition stages between the idle and flapping conditions. Note that the average values of the idle conditions were also used as a reference to obtain the cycle-average lift. This method was also used in our previous work [23]. For each set, we repeated the measurement seven times to increase the accuracy of the force data. The resultant force was obtained by averaging the results of these measurements.

To acquire the input power, we used a four-channel oscilloscope (TDS 2024, Tektronix Inc.) and two 1 Ohm resistors ($R = 2$ Ω, tolerance = 1%) (Fig. 4B). The oscilloscope records the output voltage from the power supply ($V_s$) and the input voltage applied to the flapping-wing system ($V_o$). Thus, $V_s - V_o$ is the voltage across the resistors. Accordingly, the current ($I$) applied to the resistors and flapping-wing system can be obtained by $I = (V_s - V_o)/R$. The input power of the flapping-wing system is achieved by multiplying the $V_o$ by the current $I$. Moreover, the flapping frequency was recorded by using a high-speed camera (Photron Ultima APX) at 2,000 fps.

### B. Results

Fig. 5 shows the experimental results of force and power for different configurations. We found that a wing with a stroke amplitude as high as 190° generates 28.9 ± 6.8% (mean ± SD) higher lift-to-power ratio than that flapping at an amplitude of about 120° (Fig. 5A). Enlarging the wing area by about 27% (from 20.1 cm$^2$ to 25.5 cm$^2$) increased the lift-to-power ratio by 10.9 ± 4.1% at the stroke amplitude of 190°. In addition, it is clear that a wing with larger wing areas and stroke amplitudes flaps at lower frequencies to produce similar lift (Fig. 5B). Furthermore, even though a larger wing draws a higher current, it requires a lower applied voltage to generate a similar lift to that produced by a smaller wing, allowing the use of a single lithium polymer battery (3.7 V) to provide sufficient lift to maintain the robot airborne (Figs. 5C and 5D). Thus, among the above tested configurations, we could obtain the best wing with a wing length of 9 cm and area of 25.5 cm$^2$ flapping at an amplitude of about 190° for economic flight.

## IV. EFFECT OF INBOARD WING SURFACE

Fig. 6 shows that during flapping motion, the wing is twisted along the wingspan (see *Appendix* for the wing kinematic analysis). With this twisted configuration, we found that the geometric angles-of-attack $α_g$ (AoA) from the wing root to about 25% wingspan, where the reinforced vein is located (Fig. 4A), are extremely high (60° < $α_g$ < 90°) (Fig. 6B), dominantly generating a high drag rather than lift. Thus, a flapping-wing robot may require high power to flap the wing with these high AoAs. Therefore, the wing area near the wing root may be useless for flapping-wing robots with twisted wing shape as found in our KUBeetle-S and other robots, such as the Nano Hummingbird. To demonstrate this hypothesis, we removed the wing membrane from the wing root to about 25% wingspan of the best wing found in the previous section (Fig. 4A). The lift and power requirements of this modified wing were then theoretically and experimentally obtained and compared to those of the original intact wing. Because the wing membrane is made of light-weight material, the masses of the two wings are almost the same. Therefore, in the theoretical unsteady blade-element theory (UBET) model (see *Appendix*), which was validated in our previous works [21], [25], only aerodynamic components were calculated. Fig. 7 shows the estimated cycle-average lifts and aerodynamic powers along the wingspan produced by the two wings. Because the two wings are the same except for the cut area, the lift and power generated by only this area are different. The results show that the modified wing produces about 2% less lift and aerodynamic power than the intact wing. Thus, the lift-to-aerodynamic-power ratios of the two wings remain the same, demonstrating that aerodynamic efficiency is not affected by the cutout area in the inboard wing.

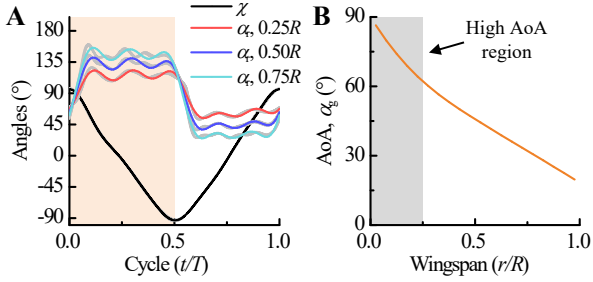

Fig. 6. (A) Time course of the stroke angle ($\chi$) and rotational angle ($\alpha_r$) at different wing sections along the wingspan. Measured data from three flapping cycles are denoted in gray color. Shaded area denotes the downstroke period. (B) Distribution of the geometric AoA ($\alpha_g$) along the wingspan.

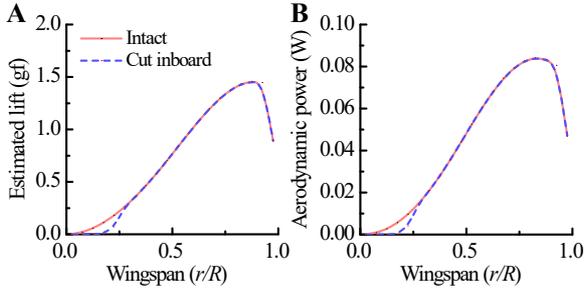

Fig. 7. Distribution of (A) lift and (B) aerodynamic power along the wingspan, produced by the intact and modified wings at a flapping frequency of 17.3 Hz.

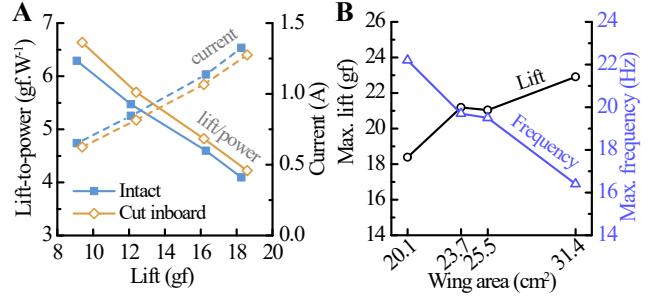

Fig. 8. (A) Effect of wing inboard area on lift-to-power ratio. (B) Maximum lift and frequency powered by a single lithium polymer battery (3.7V) for different wing areas at an amplitude of about 190°. Maximum force SD = ±0.2 gf).

TABLE I. MASS BREAKDOWN OF THE ROBOT

| Components | Mass (g) | Percentage (%) |
|---|---|---|
| Flapping mechanism | 2.6 | 16.5 |
| Driving motor | 3.5 | 22.2 |
| Servos | 1.8 | 11.4 |
| Control linkages | 0.7 | 4.4 |
| Control boards | 1.6 | 10.1 |
| Wiring and supporting frames | 1.2 | 7.6 |
| Wings | 0.4 | 2.5 |
| Battery (160 mAh) | 4.0 | 25.3 |
| **Total mass** | **15.8** | **100** |

However, the experiment shows that less inboard wing area improves the lift-to-power ratio to 4.8 ± 1.2%, while insignificantly affecting the cycle-average lift generation (Fig. 8A), i.e., to produce the same amount of lift, the modified wing requires lower input power than the intact wing. To investigate the effect of the inboard wing area on the power requirement, we decomposed the total input power ($P_{in}^f$) into power components as follows:

$$P_{in}^f = P_{loss}^f + P_{fm}^f + P_{aero}^f + P_i^f, \quad (3)$$

where, $P_{loss}^f$, $P_{fm}^f$, $P_{aero}^f$, and $P_i^f$ denote the Joule loss in the motor, power spent by the flapping mechanism only, aerodynamic power, and inertial power, respectively. We assume that the $P_i^f$ values of the two wings are identical due to them having the same wing mass. In addition, the two wings flap at a similar frequency to produce the same lift, resulting in only a small difference in $P_{fm}^f$. From the above theoretical results by the UBET model, the difference in $P_{aero}^f$ is only about 2%. Thus, the remaining $P_{loss}^f$ is also affected by the cutout area, which can be estimated by $P_{loss}^f = R_m I^2$, where $R_m$ is the motor resistance. By analyzing the current, we found that the modified wing draws lower current than the intact wing (Fig. 8A), resulting in lower power loss $P_{loss}^f$ and therefore lower total input power $P_{in}^f$. The robot employing the modified wing with a total area of 23.7 cm² produced a peak lift-to-weight ratio of 1.34 (body mass was 15.8 g) at 3.7 V application of a single battery (Fig. 8B), which is within the operating range of the DC motor.

## V. FLIGHT EXPERIMENTS

Before the free flight tests, tethered flight tests were used to trim the robot in its bias roll and pitch torques, which are generated by the asymmetric flapping motions caused by imperfect fabrication. The process was first carried out by adjusting the location of the battery to relocate the robot's center of mass. However, this method could not perfectly eliminate all the bias torques. Therefore, we added offset torques (by adjusting the potentiometers in the remote control transmitter) and performed tethered flight tests until the robot took off vertically, as a result of no bias torque generation. We then performed indoor free flight tests in a 3 m × 3 m × 3 m room with no wind conditions. For endurance and outdoor flight tests, we used a camera (Nikon D7000, 24 fps) to record the flight. Meanwhile, we used a high-speed camera (Photron Ultima APX, 500 fps) to obtain the flapping frequency during hover, and to analyze flight speeds during takeoff,

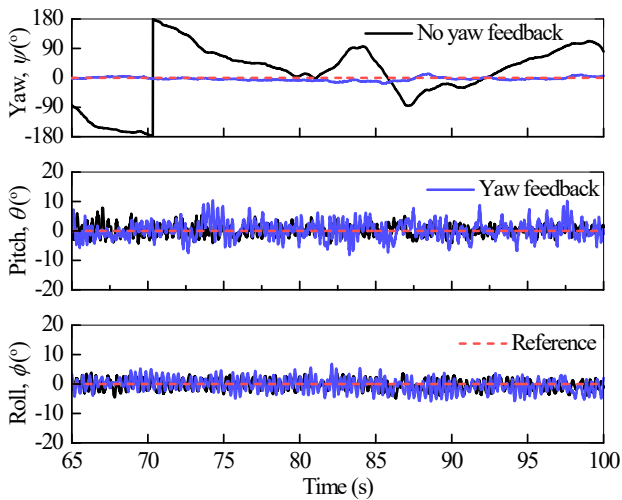

Fig. 9. Attitude performance of the KUBeetle-S robot during hovering flights with and without heading feedback control.

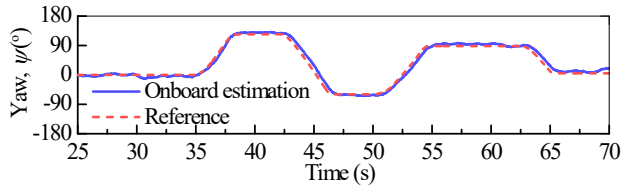

Fig. 10. Response of the heading orientation with respect to the reference of yaw input command.

forward/backward, and sideways flights. The robot was set to fly across the camera's field of view.

The final KUBeetle-S prototype with a wingspan of 20 cm weighs 15.8 g (no legs, Table 1). The wing loading, which is the ratio between body weight and total wing area, is about 37.5 N/m$^2$. Even though it is smaller than the wing loading of the Nano Hummingbird (~ 45.7 N/m$^2$), it is similar to that of a real beetle (38.9 ± 3.7 N/m$^2$) [25]. Fig. 9 shows body attitude angles of the robot during hovering flight tests. Without yaw feedback control, although upright stability (represented by roll and pitch angles) was maintained, the robot experienced rotation around the vertical axis (Fig. 9, black solid lines). However, when yaw feedback control was activated, the yaw angle as well as roll and pitch angles (blue solid lines) showed small variations from the reference of 0° (red dashed line). In addition, the heading orientation responded well to the yaw command input (Fig. 10), demonstrating the effectiveness of the proposed control algorithm using only gyroscope signals. Additionally, the effective control torque generation allows the KUBeetle-S to hover (Fig. 11A) and perform maneuvers in any direction with a takeoff speed of about 1.1 m/s, and quick transition from hover to forward/backward and sideways flights at about 2.5 m/s (Fig. 11, B and C, Movie S1). The robot could also fly outdoors in a relatively low wind condition of less than 1 m/s (Fig. 11D and Movie S2). The endurance test showed that the robot could hover for 8.8 minutes (Movie S3) by flapping its wings at a frequency of around 18 Hz. This endurance is longer than those of the similar-size tailless robots, except for the Nano Hummingbird (11 minutes, 17.5 g Saturn version) (Fig. 12). To demonstrate the capability for real applications, we added a 2.3 g payload (2 g dummy weight + 0.3 g legs), which was the estimated mass of a micro vision system, to the robot. At a body mass of 18.1 g, the robot could remain stable for 7.5 minutes in the air, which is longer than the endurance of the Nano Hummingbird with onboard payload. Note that the battery size used in the 19 g Nano Hummingbird was reduced from that in the 17.5 g Saturn version to fit into the body shell [9]. Thus, our two-winged robot demonstrated its ability to enter the next phase of autonomous flight.

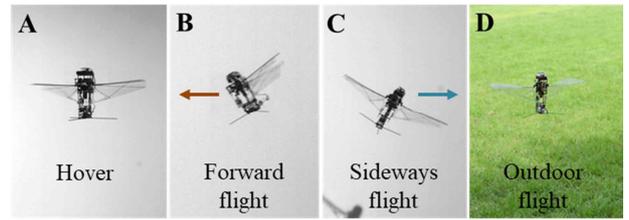

Fig. 11. High-speed camera images of the robot during indoor (A) hovering, (B) forward, and (C) sideways flights. (D) Snapshot image of the robot in outdoor flight.

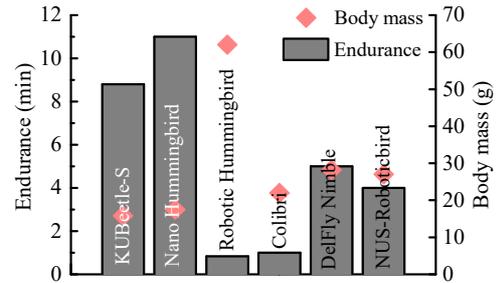

Fig. 12. Maximum flight endurance of the KUBeetle-S in comparison to those of the similar-size tailless robots (Nano Hummingbird [9], Colibri [11], DelFly Nimble [13], NUS-Roboticbird [14], and Robotic hummingbird [26].

## VI. DISCUSSION

Weighing only 15.8 g, the KUBeetle-S is currently the lightest tailless two-winged robot that can perform free controlled flight using onboard power source and electronics. Using a larger wing with less inboard area, we showed that the robot improves lift and lift-to-power ratio. In addition, the operating voltage of the robot is also lower, allowing the use of a 3.7 V battery as a power source to avoid overheating of the driving motor in long period of operation. As a result, the robot with only 3.7 % weight reduction improved flight time by about 193% (8.8 min vs. 3 min), compared to our previous 16.4 g robot version driven by 7.4 V battery [20].

A previous study indicated that the flapping wing is more aerodynamically efficient than the rotary wing [27]. However, unlike rotorcraft, most of the motor-driven flapping-wing robots use transmission systems that consume power (frictional loss) to convert rotary motion of the driving motor to the proper flapping motions of wings. Moreover, due to oscillatory wing motion, additional energy (inertial loss) is required to overcome inertial force and accelerate the wings. Despite consuming lower aerodynamic power, the flapping-wing robot incorporated with a transmission linkage may suffer from the abovementioned energy losses to compete with a similar-scale rotorcraft. For example, the 33 g Black Hornet Nano rotorcraft can sustain a flight of up to 25 minutes, which is longer than the 11 minute flight of the Saturn [9]. Therefore, alternative methods in designing a lightweight insect-inspired motor-driven robot are required for it to last longer in the air. A direct-driven robot with energy-recovery elastic elements eliminating frictional and inertial losses may be regarded as a promising candidate [19]. Other than that, flapping-wing robots may show better flight agility [13], and with low-frequency wing motion be more human-friendly than the rotorcraft. In addition, they can enable the secrets of the agile maneuvers of insects to be discovered, such as quick banked turns in flies [13]. The robots can also be potentially disguised as spy insects to explore the life of real insects in the forest or used in defense and military applications to conduct secret missions. However, they should be improved to show more capabilities to be ready for real applications, such as low-noise flight, autonomous flight, swarm flight, obstacle avoidance and collision resistance in confined spaces, flight in various weather conditions, and the performance of multiple locomotion modes.

## VII. CONCLUSION

This letter reports the achievement in flight endurance of our 15.8 g KUBeetle-S robot by using a low-voltage power source. The performance tests show that a larger wing with an area of 23.7 cm$^2$ and less inboard wing area improves lift-to-power ratio by about 15.9 ± 4.9%. We also prove that a higher stroke amplitude of flapping wings is more beneficial for economical flight. By using a single lithium polymer battery, the robot produced a peak lift-to-weight ratio of 1.34 and hovered for 8.8 minutes. By adding payload simulating the mass of a vision system, the 18.1 g robot could remain stable for 7.5 minutes in the air. Moreover, the robot could perform a takeoff at 1.1 m/s, and quick transitions from hovering to forward, backward, and sideway flights at about 2.5 m/s. It could also fly outdoors in relatively low wind conditions, demonstrating the capability for outdoor as well as indoor applications. Future work will focus on further extension of the flight time, and installation of an onboard vision system for flight navigation. Improvement of the stability during flight transition, such as from quick forward flight to hover, and implementation of altitude hold mode, will also be considered for better flight agility of the robot.

## VIII. APPENDIX

### A. Wing kinematics

We used three synchronized high-speed cameras (Photron Ultima APX, 2,000 fps, 1,024 × 1,024 pixels) to record the flapping wing motion, and digitized the stroke angle ($\chi$) and rotational angle ($\alpha_r$), which is the angle between the upstroke moving direction and the wing section, using DLT software [28]. More details on the experiment can be found in [20], [22]. The measured angles were then fitted to use as inputs of the theoretical model for force and power estimation. The fitted function is described as follows:

$$\Lambda(t) = a_0 + \sum_{n=1}^{5}[a_n \cos(2n\pi ft) + b_n \sin(2n\pi ft)], \quad (4)$$

where, $a_0$, $a_n$ and $b_n$ are the fitted coefficients, $\Lambda(t)$ is either the stroke angle or rotational angle, and $f$ is the flapping frequency.

### B. Aerodynamic force estimation

We divided the wing into 20 spanwise sections, and used the unsteady blade-element theory model [25, 29] to calculate three aerodynamic force components (translational force ($dF_T$), added mass force ($dF_A$), and rotational force ($dF_R$)) at each wing section $dr$. Vertical force or lift in the $\zeta$-direction, which is perpendicular to the stroke plane, and horizontal force in the $\eta$-direction, which is tangential to the moving direction of the section, produced in a wing section $dr$ and an instant time $t$ can be described as the following equations:

$$dF_\eta(r,t) = dF_{T\eta}(r,t) + dF_{A\eta}(r,t) + dF_{R\eta}(r,t), \quad (5)$$

$$dF_\zeta(r,t) = dF_{T\zeta}(r,t) + dF_{A\zeta}(r,t) + dF_{R\zeta}(r,t), \quad (6)$$

where,

$$dF_{T\eta}(r,t) = -\frac{1}{2}\rho c_r (C_L \sin\varphi + C_D \cos\varphi)(V_T^2 + V_i^2)dr, \quad (7)$$

$$dF_{A\eta}(r,t) = \frac{\pi}{4}\rho c_r^2 a_w \sin\alpha_g \sin\alpha_r dr, \quad (8)$$

$$dF_{R\eta}(r,t) = -\rho V_T c_{rot} \dot{\alpha}_r c_r^2 \sin\alpha_r dr, \quad (9)$$

$$dF_{T\zeta}(r,t) = \frac{1}{2}\rho c_r (C_L \cos\varphi - C_D \sin\varphi)(V_T^2 + V_i^2)dr, \quad (10)$$

$$dF_{A\zeta}(r,t) = \frac{\pi}{4}\rho c_r^2 a_w \sin\alpha_g \cos\alpha_r dr, \quad (11)$$

$$dF_{R\zeta}(r,t) = \rho V_T c_{rot} \dot{\alpha}_r c_r^2 \cos\alpha_r dr. \quad (12)$$

In the above equations, $\rho$, $c_r$, $\varphi$, $V_i$, $V_T$, and $c_{rot}$ are the air density, wing chord length, induced angle, induced velocity, translational velocity, and rotational coefficient $c_{rot} = \pi\left(0.75 - l_r/c_r\right)$ [30], respectively. By a combination of the momentum and the blade element theories to solve for the translational vertical force $dF_{T\zeta}$, the induced velocity $V_i$ could be obtained. The acceleration $a_w$ of the wing section $dr$ can be described as follows [29]:

$$a_w = [r\ddot{\chi} + (c_r/2 - l_r)\dot{\chi}^2 \cos\alpha_r]\sin\alpha_r + (c_r/2 - l_r)\ddot{\alpha}_r, \quad (13)$$

where, $r$ is the distance from the section $dr$ to the flapping axis, and $l_r$ is the chordwise distance from the leading edge and the rotational axis. The lift and drag coefficients ($C_L$ and $C_D$) are functions of the effective AoA $\alpha_e$ and Reynolds number $Re$ as given below [31]:

$$C_L = \left(1.966 - 3.94Re^{-0.429}\right)\sin 2\alpha_e, \quad (14)$$

$$C_D = 0.031 + 10.48Re^{-0.764} + \left(1.873 - 3.14Re^{-0.369}\right)(1 - \cos 2\alpha_e), \quad (15)$$

where, $Re = \dfrac{2\bar{c}\Phi fR}{\upsilon}$, $\bar{c}$ is the mean wing chord, $\Phi$ is the stroke amplitude, $f$ is the flapping frequency, $R$ is the wing length, and $\upsilon$ is the kinematic viscosity of air.

From the aerodynamic horizontal force ($dF_\eta$) in (5), the aerodynamic power ($dP_{aero}$) required to flap a wing section $dr$ at a time $t$ is estimated as follows:

$$dP_{aero}(r,t) = \dot{\chi} \cdot \left( \vec{r} \times \overrightarrow{dF_\eta}(r,t) \right). \quad (16)$$

By integrating (5), (6) and (16), the aerodynamic force and power generated by the wing in a flapping cycle can be obtained. The UBET model was validated in our previous works to provide reasonable predictions of force and power [23], [25]. In addition, in this study, the estimated lift is equivalent to about 90% of that obtained by measurement. This result is reasonable because the clap-and-fling [22] presented in the experiment was not modeled on the theoretical model.